\documentclass[conference]{IEEEtran}
\usepackage{amsmath,amssymb,amsfonts}
\usepackage{graphicx}
\usepackage{textcomp}
\usepackage{xcolor}
\usepackage{booktabs}
\usepackage{hyperref}
\usepackage{cleveref}
\usepackage[table]{xcolor}
\usepackage{multirow}
\usepackage{caption}
\usepackage{subcaption}
\usepackage{hyperref}
\usepackage{cleveref}
\usepackage{bm}
\usepackage{tikz}
\usepackage{enumitem}
\usepackage{makecell}
\usepackage[table]{xcolor}
\usetikzlibrary{arrows.meta,positioning,decorations.pathreplacing,calc,patterns,shapes.geometric,backgrounds,fit}

\newcommand{\uatom}{\textsc{UA-ToM}}

\begin{document}

\title{Belief Dynamics for Detecting Behavioral Shifts in Safe Collaborative Manipulation}
\author{
\IEEEauthorblockN{Devashri Naik, Divake Kumar, Nastaran Darabi, Amit Ranjan Trivedi}
\IEEEauthorblockA{
University of Illinois Chicago\\
Email: \{dnaik6@uic.edu, dkumar33@uic.edu, ndarab2@uic.edu, amitrt@uic.edu\}}
}
\maketitle

\begin{abstract}
Robots operating in shared workspaces must maintain safe coordination
with other agents whose behavior may change during task execution.
When a collaborating agent switches strategy mid-episode, a robot
that continues acting under outdated assumptions can generate unsafe
motions and increased collision risk. Reliable detection of such
behavioral regime changes is therefore critical for safe collaborative
manipulation. We study regime-switch detection under controlled
non-stationarity in shared-workspace manipulation tasks in ManiSkill.
Across ten detection methods and five random seeds, enabling
detection reduces post-switch collisions by 52\%. However, mean
performance hides severe reliability failures: under operationally
realistic tolerance ($\pm 3$ steps), methods separate into three
tiers ranging from 86\% to 30\% detection, a difference invisible
at the standard $\pm 5$ window where all methods achieve 100\%.
We introduce \uatom{}, a lightweight belief-tracking module that
augments frozen vision-language-action (VLA) control backbones
using selective state-space dynamics, causal attention, and
prediction-error signals. Across five seeds and 1200 episodes,
\uatom{} achieves the highest detection rate among unassisted
methods (85.7\% at $\pm 3$) and the lowest close-range time
(4.8 steps in the danger zone), including lower than an Oracle
with perfect detection (5.3 steps). Analysis shows that
hidden-state update magnitude increases by $17\times$ at regime
switches and decays over roughly 10 timesteps while the
discretization step converges to a near-constant value
($\Delta_t \approx 0.78$), indicating that detection sensitivity
arises from learned state dynamics rather than input-dependent
gating. Cross-domain experiments in Overcooked reveal complementary
architectural roles where causal attention dominates detection in
discrete coordination while prediction-error signals dominate in
continuous manipulation. \uatom{} introduces 7.4\,ms inference
overhead, corresponding to 14.8\% of a 50\,ms control budget,
enabling reliable regime-switch detection for collaborative
manipulation without modifying the base policy.
\end{abstract}

\section{Introduction}
\label{sec:intro}

\begin{figure}[t]
\centering
\includegraphics[width=\columnwidth, height = 6cm]{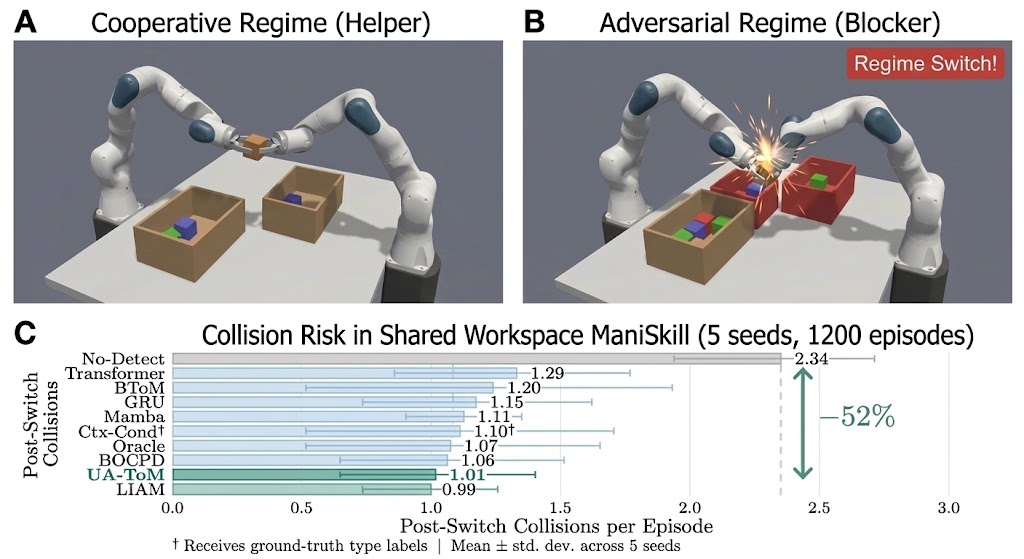}
\caption{Shared-workspace regime-switch detection. \textbf{Top}: a
collaborating agent switches from cooperative (Helper) to adversarial
(Blocker), creating collision risk. \textbf{Bottom}: enabling
detection reduces post-switch collisions by 52\% across all methods
(No-Detect: $2.34$ vs.\ detection mean: ${\sim}1.1$). Among
detectors, collision rates are comparable (range 0.99--1.29),
motivating evaluation of detection reliability and adaptation quality.}
\label{fig:teaser}
\end{figure}

Robotic systems operating in shared workspaces increasingly interact
with other decision-making agents, including human operators and
autonomous robots. In such multi-agent settings, the closed-loop
behavior of one agent depends critically on the policy executed by
the other. A key challenge arises when the collaborating agent
changes its behavioral regime during task execution. If the ego
agent continues acting under outdated assumptions about the other
agent's strategy, coordination degrades and safety risk increases.

We study the problem of detecting mid-episode behavioral regime
switches in shared-workspace manipulation. The collaborating agent
follows a latent policy type that may change at an unknown time.
The ego agent observes joint state information but receives no
explicit signal when the regime changes. Reliable detection of
these switches is therefore necessary for safe adaptive
coordination.

Existing work only partially addresses this setting. Theory-of-mind
and opponent-modeling approaches~\cite{rabinowitz2018tomnet,papoudakis2021liam}
typically assume stationary behavior at evaluation time. Classical
changepoint detection~\cite{adams2007bocpd} provides statistical tools
for regime inference but is rarely studied within high-dimensional
learned control loops. Recent large-scale control
models~\cite{brohan2023rt2,kim2024openvla} focus on single-agent
manipulation and do not maintain persistent structured beliefs over
the behavior of collaborating agents. Across these strands,
evaluation commonly reports mean detection accuracy under a single
random seed and implicitly treats average performance as
representative of deployment reliability. In safety-critical
collaborative environments this assumption is problematic.
Worst-case reliability, rather than mean accuracy, determines the
safety outcome.

We evaluate regime-switch detection under controlled
non-stationarity in shared-workspace manipulation (Fig.~\ref{fig:teaser}).
Across ten methods and five random seeds in ManiSkill, enabling
detection reduces post-switch collisions by 52\% relative to a
no-detection baseline. Reliability, however, varies substantially.
Under operationally realistic detection tolerance ($\pm 3$ steps,
corresponding to 150\,ms in a 50\,ms control loop), methods
separate into three tiers: fast detectors maintain $>$82\%
detection, agent-modeling methods achieve 58--62\%, and standard
sequence models collapse to $<$34\%. At the conventional $\pm 5$
tolerance, all methods achieve 100\% and appear equivalent,
masking these differences entirely. To address this reliability gap
we introduce \uatom{}, a lightweight belief-tracking module that
augments a frozen control backbone with structured temporal
reasoning (Fig.~\ref{fig:architecture}). The module integrates
selective state-space dynamics, causal attention, and
prediction-error signals to maintain a persistent estimate of the
collaborating agent's behavioral regime. Among unassisted methods,
\uatom{} achieves the highest detection rate at $\pm 3$ (85.7\%)
and the lowest close-range time under extended episodes (4.8 steps),
including lower than an Oracle with perfect detection.

Our contributions are as follows:
\begin{itemize}
\item We show that regime-switch detection reduces post-switch
  collisions by 52\%, while standard evaluation at $\pm 5$
  tolerance masks a three-tier separation in detection speed
  visible at $\pm 3$, revealing reliability differences exceeding
  50 percentage points across methods.
\item We introduce \uatom{}, a 992K-parameter belief-tracking module
  that augments frozen control backbones and achieves the highest
  detection rate among unassisted methods while producing the lowest
  close-range time, including lower than Oracle, through smooth
  belief revision.
\item Mechanistic analysis shows robustness arises from selective
  state-space dynamics. Hidden-state update magnitude increases
  $17\times$ at regime switches, driven by a $2.3\times$ spike in
  action prediction error, while the discretization step converges
  to a near-constant value ($\Delta_t \approx 0.78$).
\item Cross-domain evaluation shows architectural contributions are
  domain-dependent. Prediction-error signals dominate detection in
  continuous manipulation while causal attention dominates in
  discrete coordination.
\end{itemize}

\section{Related Work}
\label{sec:related}

\textbf{Vision-language-action models.}
RT-2~\cite{brohan2023rt2} shows that vision-language pretraining
can transfer to robotic control through action token prediction.
OpenVLA~\cite{kim2024openvla} provides an open-source 7B-parameter
VLA trained on the Open-X Embodiment dataset~\cite{openx2024}.
Octo~\cite{octo} targets generalist robot control using a smaller
transformer backbone. These models largely map observations to
actions at each timestep and do not maintain persistent structured
beliefs about the behavior of other agents. Our approach augments
a frozen VLA backbone with an explicit belief-tracking module
without base policy changes.

\textbf{Theory of mind for multi-agent systems.}
ToMnet~\cite{rabinowitz2018tomnet} predicts agent behavior using
learned character and mental state representations. Bayesian Theory
of Mind~\cite{baker2017btom} models agents through inverse planning
under approximate rationality. LIAM~\cite{papoudakis2021liam}
infers latent intent under partial observability using attention
mechanisms. He et al.~\cite{he2016opponent} learn
opponent-conditioned policies from trajectory history. These
approaches are typically evaluated under stationary behavior or
single-seed protocols. In contrast, we study reliability under
controlled non-stationarity and show that the detection floor,
rather than mean accuracy, differentiates methods.

\textbf{Changepoint detection.}
Bayesian Online Changepoint Detection
(BOCPD)~\cite{adams2007bocpd} maintains a posterior over run length
to detect distributional shifts. Extensions incorporate Gaussian
processes~\cite{saatci2010gp} and neural density
estimators~\cite{chang2019neural}. We show that representational
separability alone is insufficient for regime-switch detection.
BOCPD achieves the highest mean Bhattacharyya distance among
evaluated models yet yields 0\% open-loop switch F1 because its
run-length posterior does not localize switch timing within the
control loop.

\textbf{Safe human-robot collaboration.}
ISO 10218~\cite{iso10218} and ISO/TS 15066~\cite{isots15066}
define safety requirements for collaborative industrial robots.
Haddadin et al.~\cite{haddadin2017} review collision detection and
reaction strategies. Predictable motion is often enforced through
smooth trajectory generation such as jerk
minimization~\cite{flash1985}. Lasota et al.~\cite{lasota2017}
survey broader approaches to safe human-robot interaction. Our work
addresses a complementary problem. We focus on detecting when a
collaborating agent changes behavioral regime during execution.

\textbf{Multi-agent reinforcement learning.}
Non-stationarity is a central challenge in multi-agent reinforcement
learning~\cite{gronauer2022,papoudakis2019}. CTDE
frameworks~\cite{lowe2017} and learned communication
protocols~\cite{foerster2016} address coordination under shared
objectives. Carroll et al.~\cite{carroll2019} show that modeling
human behavior improves coordination, while Strouse
et al.~\cite{strouse2021} study zero-shot coordination with novel
co-agents. Our focus differs in isolating regime-switch detection
from the coordination policy itself.

\section{Problem Formulation}
\label{sec:formulation}

We consider a shared-workspace manipulation episode in which an ego
robot must remain safe while another agent may change its behavior
during execution.

\textbf{Setting.} Two agents share a workspace. The ego agent
executes policy $\pi_{\text{ego}}$. The collaborating agent follows
a latent behavioral type $\theta_t \in \Theta = \{\theta_1, \ldots,
\theta_K\}$ that can switch once at an unknown time $t_s$. At each
timestep $t$, the ego observes $\mathbf{o}_t \in \mathbb{R}^d$
consisting of its own state, the observable state of the
collaborating agent, and shared object state.

\textbf{Inference tasks.} Given observation history
$\mathbf{o}_{1:t}$, the ego maintains three coupled estimates:
\begin{align}
\hat{\theta}_t &= \arg\max_\theta \, p(\theta_t \mid \mathbf{o}_{1:t})
  & \text{(type inference)} \label{eq:type} \\
\hat{s}_t &= p(t_s{=}t \mid \mathbf{o}_{1:t})
  & \text{(switch probability)} \label{eq:switch} \\
\hat{\mathbf{a}}^c_t &= \mathbb{E}[\mathbf{a}^c_t \mid \mathbf{o}_{1:t}]
  & \text{(co-agent action prediction)} \label{eq:action}
\end{align}
where $\mathbf{a}^c_t$ denotes the collaborating agent action. A
hard detection event occurs when $\hat{s}_t$ crosses a threshold
within a tolerance window around the true $t_s$.

\textbf{Relation to changepoint detection.}
BOCPD~\cite{adams2007bocpd} detects distribution shifts by
maintaining a posterior over run length $r_t$ through a hazard
function $H(\tau)$. This mechanism is well suited to identifying
that a change has occurred but is not designed to produce a
low-latency, within-episode switch signal inside a learned control
loop. Our design instead uses a selective state-space belief update
whose discretization $\Delta_t$ converges during training to a
near-constant value, so sensitivity is governed by the learned
dynamics matrices $\mathbf{A}$ and $\mathbf{B}$.

\textbf{Reliability under deployment variation.} Safety depends on
worst-case detection behavior across plausible operating conditions.
We quantify reliability of method $\mathcal{M}$ across conditions
$\mathcal{C}$ as:
\begin{equation}
R(\mathcal{M}) = \frac{\min_{c \in \mathcal{C}} D_c(\mathcal{M})}
                      {\max_{c \in \mathcal{C}} D_c(\mathcal{M})}
\label{eq:reliability}
\end{equation}
where $D_c(\mathcal{M})$ is the hard detection rate under condition
$c$. $R{=}1$ indicates identical performance across conditions.

\section{\uatom{} Architecture}
\label{sec:method}

\begin{figure*}[t]
\centering
\includegraphics[width=0.8\textwidth]{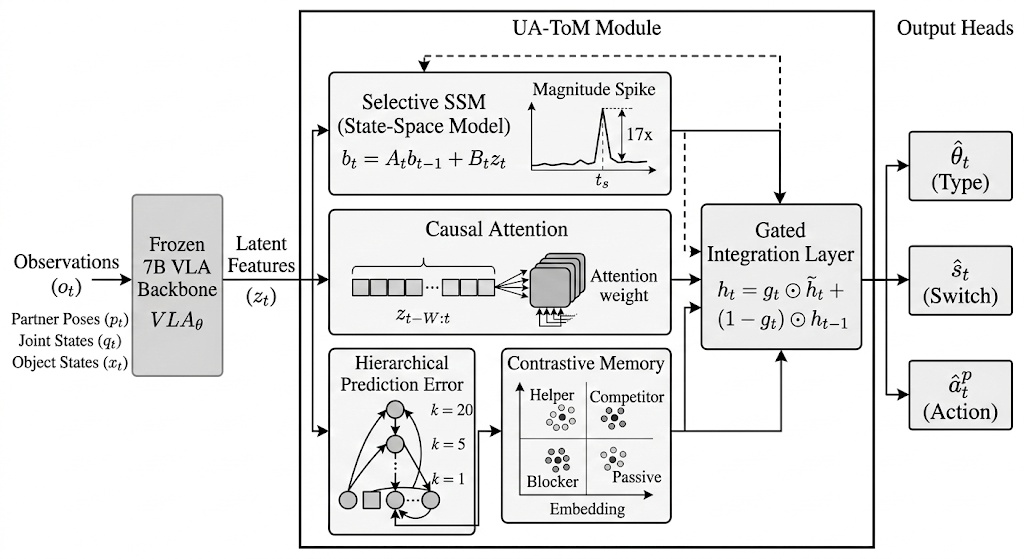}
\caption{\uatom{} computation flow. A frozen VLA backbone extracts
features $\mathbf{z}_t$. The \uatom{} module (992K parameters,
0.014\% of backbone) processes these through four pathways: a
selective SSM maintains persistent belief $\mathbf{b}_t$ via learned
dynamics $\mathbf{A}, \mathbf{B}$ (discretization $\Delta_t$
converges to 0.78); causal attention re-anchors belief to recent
observations; prediction error $e_t$ provides an explicit change
signal; and contrastive memory supplies categorical type cues via
EMA prototypes. A gated fusion layer integrates all signals to
produce type $\hat{\theta}_t$, switch probability $\hat{s}_t$, and
predicted action $\hat{\mathbf{a}}^c_t$. Internal dynamics around
the regime switch are shown in Fig.~\ref{fig:dynamics}.}
\label{fig:architecture}
\end{figure*}

\uatom{} augments a frozen VLA control backbone with a lightweight
belief-tracking module (Fig.~\ref{fig:architecture}). The backbone
maps observations to latent features
$\mathbf{z}_t \in \mathbb{R}^{d_z}$. \uatom{} maintains a
persistent belief state that summarizes the recent interaction
history and produces two outputs, a collaborating-agent type
estimate and a regime-switch signal. The module has 992K parameters,
about 0.014\% of a 7B backbone.

\textbf{Design overview.} Regime switches are rare and
safety-critical, so the detector must be stable under steady
interaction and highly sensitive at behavioral changes. \uatom{}
therefore combines four signals: persistent belief dynamics to
accumulate evidence over time while suppressing steady-state noise;
short-horizon re-anchoring to prevent belief drift under partial
observability; prediction-error evidence to create a direct change
signal tied to action mismatches; and type prototypes to provide a
categorical cue that complements continuous belief dynamics.

\textbf{Selective state-space belief update.} The core belief state
$\mathbf{b}_t$ evolves through a selective state-space update:
\begin{align}
\mathbf{b}_t &= \bar{\mathbf{A}}_t \mathbf{b}_{t-1}
              + \bar{\mathbf{B}}_t \mathbf{z}_t \label{eq:ssm} \\
\boldsymbol{\Delta}_t &= \mathrm{softplus}(\mathbf{W}_\Delta
  \mathbf{z}_t + \mathbf{d}) \label{eq:delta}
\end{align}
with $\bar{\mathbf{A}}_t = \exp(\boldsymbol{\Delta}_t \odot
\mathbf{A})$ and $\bar{\mathbf{B}}_t = \boldsymbol{\Delta}_t \odot
\mathbf{B}$. Although $\boldsymbol{\Delta}_t$ is input dependent in
form, it converges during training to a near-constant value
(approximately 0.78), which stabilizes the effective update rate.
Sensitivity to regime changes is therefore carried by the learned
dynamics matrices $\mathbf{A}$ and $\mathbf{B}$.

\textbf{Short-horizon re-anchoring with causal attention.}
\begin{equation}
\mathbf{c}_t = \mathrm{CausalAttn}(\mathbf{Q}{=}\mathbf{b}_t,\;
\mathbf{K}{=}\mathbf{z}_{t-W:t},\; \mathbf{V}{=}\mathbf{z}_{t-W:t})
\end{equation}
This provides a complementary pathway that can recover from stale
belief when per-step evidence is sparse, which is particularly
important in discrete coordination tasks.

\textbf{Prediction-error evidence.}
\begin{equation}
e_t = \|\hat{\mathbf{a}}^c_t - \mathbf{a}^c_t\|
\end{equation}
During training we also regularize the representation using
multi-scale averages of $e_t$ over $k \in \{1, 5, 20\}$, but at
inference we use only the single-step error to avoid additional
temporal computation.

\textbf{Prototype memory for type cues.} We maintain exponential
moving-average prototypes $\mathbf{m}_\theta$ for each behavioral
type:
\begin{equation}
\mathbf{m}_\theta \leftarrow \alpha \mathbf{m}_\theta
  + (1 - \alpha) \mathbf{b}_t, \quad \text{if } \theta_t = \theta
\end{equation}
and compute cosine similarities $\mathrm{sim}_\theta =
(\mathbf{b}_t \cdot \mathbf{m}_\theta) /
(\|\mathbf{b}_t\| \|\mathbf{m}_\theta\|)$.

\textbf{Evidence fusion and outputs.}
\begin{align}
\mathbf{h}_t &= g_t \odot \tilde{\mathbf{h}}_t
              + (1 - g_t) \odot \mathbf{h}_{t-1} \\
g_t &= \sigma(\mathbf{W}_g[\mathbf{b}_t;\, \mathbf{c}_t;\, e_t;\,
  \mathrm{sim}])
\end{align}
From $\mathbf{h}_t$, the module predicts collaborating-agent type
$\hat{\theta}_t$ and switch probability $\hat{s}_t$.

\textbf{Training objective.}
$\mathcal{L} = \mathcal{L}_{\text{act}} + \lambda_\theta
\mathcal{L}_{\text{type}} + \lambda_s \mathcal{L}_{\text{sw}}$,
where $\mathcal{L}_{\text{act}} = \mathrm{MSE}(\hat{\mathbf{a}}^c_t,
\mathbf{a}^c_t)$, $\mathcal{L}_{\text{type}} =
\mathrm{CE}(\hat{\theta}_t, \theta_t)$,
$\mathcal{L}_{\text{sw}} = \mathrm{CE}(\hat{s}_t, s_t)$, and we
set $\lambda_\theta{=}1$ and $\lambda_s{=}2$ to mitigate switch
class imbalance. All learned baselines use the same objective and
training data for controlled comparison. We optimize with Adam
(lr $10^{-3}$, weight decay $10^{-4}$), batch size 64, for
50 epochs over 12 transition types, while keeping the backbone
frozen. Training takes about 7 minutes on an RTX 6000 Ada GPU.

\section{Experimental Setup}
\label{sec:setup}

\textbf{ManiSkill shared workspace.} We evaluate regime-switch
detection in a shared-workspace manipulation setting using two
7-DOF Franka Panda arms operating in a
0.6\,m $\times$ 0.4\,m workspace. The ego arm performs
pick-and-place while the collaborating agent follows one of four
behavioral types:
\begin{itemize}
\item \textbf{Helper}: yields workspace and moves the object toward ego.
\item \textbf{Competitor}: races to grasp and retreats on success.
\item \textbf{Blocker}: interposes between the ego and the target.
\item \textbf{Passive}: performs minimal motion.
\end{itemize}
The collaborating agent switches type at
$t_s \sim \mathcal{U}[30, 100]$ within 200-step episodes.
Observations are 48-dimensional and include tool-center-point poses
(14D), joint positions (18D), object state (7D), relative positions
(6D), inter-effector distance (1D), and direction vector (2D). We
evaluate 240 episodes per seed across 12 transition types, totaling
1200 episodes over five seeds. For extended evaluation, we run
300-step episodes across ten seeds using the same partner types and
transition structure.

\textbf{Overcooked cross-domain evaluation.} We additionally
evaluate in the Overcooked environment~\cite{carroll2019}, where
the collaborating agent follows one of four behavioral types:
\textbf{Reliable}, \textbf{Lazy}, \textbf{Saboteur}, and
\textbf{Erratic}. Experiments are conducted across three layouts
(Cramped Room, Asymmetric Advantages, Counter Circuit). All agent
types follow identical cooperative behavior during an initial
warm-up period ($t_{\text{div}} \sim \mathcal{U}[30, 50]$) before
diverging. Observations are 192-dimensional feature vectors.

\textbf{Baselines.} We compare against nine methods spanning four
categories. Sequence models include GRU (two layers, 128 hidden
units), Transformer (four layers, four heads, 64-dimensional
embeddings), and Mamba~\cite{gu2023mamba}. Agent-modeling
approaches include LIAM~\cite{papoudakis2021liam} and
BToM~\cite{baker2017btom}, implemented with a particle filter
using 100 particles. As a classical baseline we evaluate
BOCPD~\cite{adams2007bocpd} using a Student-$t$ observation model
applied to GRU prediction errors with fixed hazard rate $H{=}1/100$.
Control conditions include Oracle (ground-truth switch timing),
Context-Conditioned (ground-truth type provided), and No-Detection
(fixed strategy without adaptation). All learned baselines use the
same multi-task objective, training data, and comparable
hyperparameter budgets.

\textbf{Metrics.} \emph{Hard detection rate} measures the fraction
of regime switches detected within a tolerance window around the
ground-truth switch. We report at two thresholds: $\pm 5$ steps
(conventional) and $\pm 3$ steps (operationally realistic,
corresponding to 150\,ms in a 50\,ms control loop).
\emph{Detection latency} measures the number of timesteps between
the true switch and the first detection. \emph{Post-switch
collisions} count end-effector proximity events ($< 0.1$\,m)
occurring after $t_s$. \emph{Close-range time (CRT)} counts
post-switch timesteps where end-effector distance is below
0.15\,m ($1.5\times$ the collision threshold), capturing sustained
proximity risk even when hard collisions are avoided. A method with
low collisions but high CRT operates consistently near the danger
boundary. Reliability is defined as
$R = \min_c D_c / \max_c D_c$ (eq.~\eqref{eq:reliability}). We
also report the coefficient of variation (CV) of $D_c$ across seeds.

\section{Results}
\label{sec:results}

\textbf{Detection reduces collisions.} We first quantify the safety
benefit of enabling regime-switch detection. Table~\ref{tab:benefit}
compares a no-detection policy against policies equipped with switch
detectors (see also Fig.~\ref{fig:teaser}, bottom). Enabling
detection reduces post-switch collisions from $2.34 \pm 0.40$ to
$1.11 \pm 0.12$ per episode, corresponding to a 52\% reduction.
This improvement occurs consistently across detection-capable
methods. Once a switch is detected, downstream adaptation produces
similar collision outcomes across detectors (range 0.99--1.29
collisions per episode). The primary safety bottleneck is therefore
not how the switch is handled after detection, but whether the switch
is detected reliably in the first place.

\begin{table}[t]
\centering
\caption{Effect of regime-switch detection on post-switch collision
rate (5 seeds). Lower is better.}
\label{tab:benefit}
\setlength{\tabcolsep}{6pt}
\begin{tabular}{@{}lc@{}}
\toprule
\textbf{Condition} & \textbf{Post-Switch Collisions} \\
\midrule
No detection & $2.34 \pm 0.40$ \\
Detection (mean across methods) & $1.11 \pm 0.12$ \\
\midrule
Reduction & 52\% \\
\bottomrule
\end{tabular}
\end{table}

\begin{figure*}[t]
\centering
\includegraphics[width=0.88\textwidth]{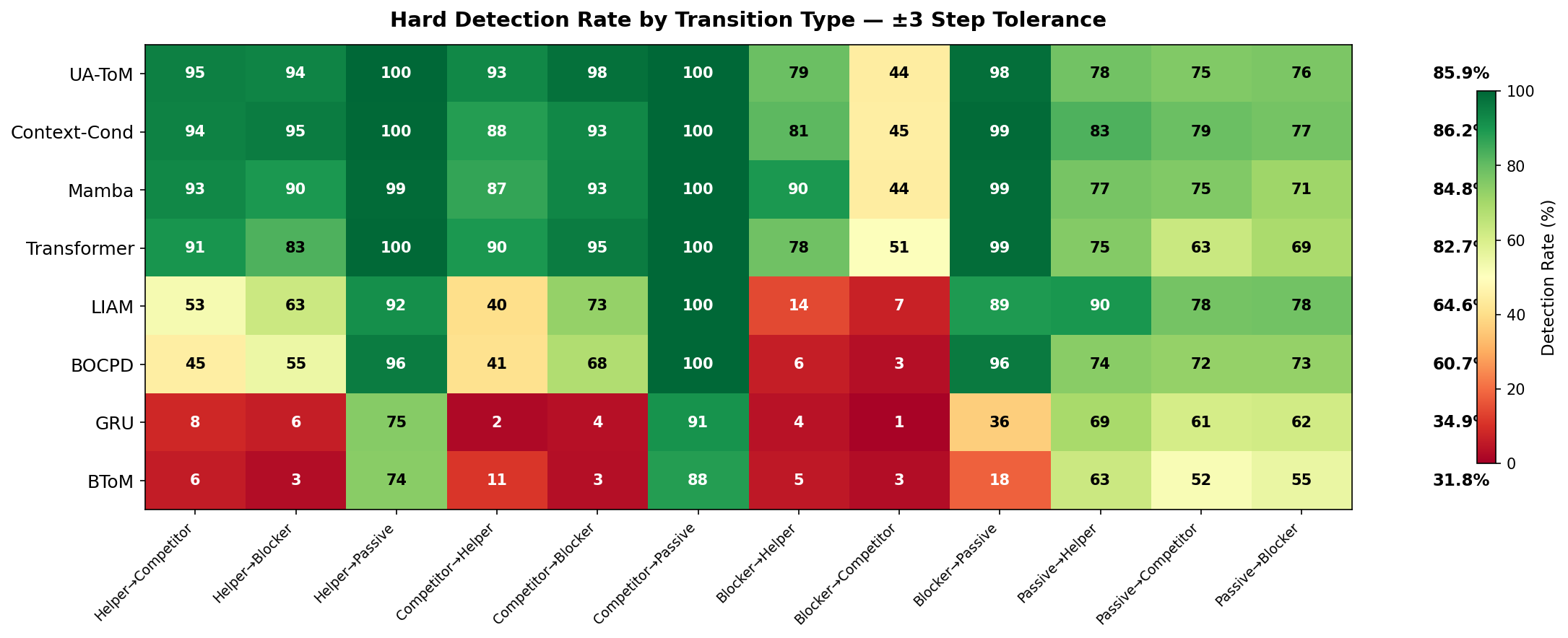}
\caption{Detection reliability heatmap at $\pm 3$ step tolerance.
Hard detection rate (\%) per method and transition type across 12
behavioral transitions (e.g.\ Helper$\to$Blocker,
Competitor$\to$Passive). Rows are sorted by mean detection rate.
Methods separate into three tiers: fast detectors ($>$82\%, blue),
agent-modeling (58--62\%, yellow), and slow recurrent models
($<$34\%, red). At the standard $\pm 5$ tolerance, all cells would
be 100\%, masking these operationally relevant differences.
Transition difficulty varies: Helper$\to$Competitor is consistently
hardest across methods due to similar spatial approach patterns.}
\label{fig:heatmap}
\end{figure*}

\textbf{Detection reliability across seeds.}
Table~\ref{tab:perseed} and Fig.~\ref{fig:heatmap} report hard
detection rates under operationally realistic tolerance ($\pm 3$
steps). At the standard $\pm 5$ tolerance, all methods achieve
100\% and appear equivalent. At $\pm 3$, methods separate into
three tiers.

\begin{table}[t]
\centering
\caption{Hard detection rate (\%) per seed at $\pm 3$ step
tolerance (150\,ms at 50\,ms control rate). At $\pm 5$, all methods
achieve 100\%. $^\dagger$Receives ground-truth type.}
\label{tab:perseed}
\setlength{\tabcolsep}{2.5pt}
\scriptsize
\begin{tabular}{@{}l|ccccc|cc@{}}
\toprule
& \multicolumn{5}{c|}{\textbf{Hard Detection (\%)}} & & \\
\textbf{Method} & \textbf{S0} & \textbf{S1} & \textbf{S2} & \textbf{S3} & \textbf{S4} & \textbf{Mean} & $R$ \\
\midrule
Ctx-Cond.$^\dagger$ & 86.8 & 87.8 & 85.5 & 87.5 & 82.8 & 86.1 & 0.94 \\
\textbf{\uatom{}} & \textbf{85.3} & \textbf{83.5} & \textbf{89.2} & \textbf{87.3} & \textbf{83.0} & \textbf{85.7} & \textbf{0.93} \\
Mamba & 85.5 & 85.0 & 87.3 & 85.2 & 81.8 & 85.0 & 0.94 \\
Transformer & 83.3 & 83.3 & 82.8 & 83.8 & 79.2 & 82.5 & 0.95 \\
\midrule
LIAM & 62.3 & 56.8 & 61.0 & 63.8 & 68.2 & 62.4 & 0.83 \\
BOCPD & 60.8 & 56.7 & 60.8 & 59.5 & 52.8 & 58.1 & 0.87 \\
\midrule
GRU & 35.7 & 37.8 & 27.3 & 31.8 & 33.3 & 33.2 & 0.72 \\
BToM & 33.8 & 30.2 & 33.8 & 29.3 & 24.7 & 30.4 & 0.73 \\
\bottomrule
\end{tabular}
\end{table}

Fast detectors with median latency $\leq$2 steps (\uatom{}, Mamba,
Transformer) maintain $>$82\% across all seeds. Agent-modeling
methods (LIAM, BOCPD) achieve 58--62\%. Standard recurrent models
(GRU, BToM) collapse to 30--33\%. The three-tier separation
reflects detection latency: \uatom{}'s median latency is 2 steps
(P75 = 3), while GRU's median is 4 (P75 = 6). Under a $\pm 3$
window, any detection arriving after step 3 is counted as a miss.

Among unassisted methods, \uatom{} achieves the highest mean
detection (85.7\%) and reliability ratio ($R = 0.93$). The
Context-Conditioned baseline (86.1\%) slightly exceeds \uatom{} but
receives ground-truth type labels, reflecting smooth representational
transitions from privileged information rather than learned temporal
reasoning.

\textbf{Aggregated closed-loop performance.}
Table~\ref{tab:full} summarizes closed-loop outcomes across all
episodes. All detection-based methods substantially outperform the
No-Detection baseline, reducing post-switch collisions from 2.34 to
approximately 0.99--1.29 per episode. Differences between detectors
are therefore modest once a switch is successfully identified.

\begin{table}[t]
\centering
\caption{Closed-loop performance over five seeds (1200 episodes).
$^\dagger$Privileged.}
\label{tab:full}
\setlength{\tabcolsep}{3.5pt}
\small
\begin{tabular}{@{}lccc@{}}
\toprule
\textbf{Method} & \textbf{Latency}$\downarrow$ & \textbf{Post-Sw.\ Coll.}$\downarrow$ & \textbf{Total Coll.}$\downarrow$ \\
\midrule
Oracle & $1.0 \pm 0.0$ & $1.07 \pm 0.58$ & $2.69 \pm 1.02$ \\
\midrule
\textbf{\uatom{}} & $\mathbf{9.9 \pm 0.8}$ & $1.01 \pm 0.39$ & $2.66 \pm 0.52$ \\
Ctx-Cond.$^\dagger$ & $9.4 \pm 1.1$ & $1.10 \pm 0.61$ & $2.64 \pm 0.73$ \\
Transformer & $10.5 \pm 1.3$ & $1.29 \pm 0.48$ & $3.35 \pm 1.10$ \\
Mamba & $10.6 \pm 1.0$ & $1.11 \pm 0.24$ & $3.20 \pm 0.51$ \\
LIAM & $11.0 \pm 0.6$ & $0.99 \pm 0.28$ & $2.46 \pm 0.36$ \\
BOCPD & $11.9 \pm 0.8$ & $1.06 \pm 0.45$ & $2.96 \pm 0.68$ \\
GRU & $12.4 \pm 0.5$ & $1.15 \pm 0.47$ & $2.62 \pm 0.41$ \\
BToM & $12.5 \pm 0.9$ & $1.20 \pm 0.73$ & $2.72 \pm 1.17$ \\
\midrule
No-Detect & $86.2 \pm 1.1$ & $2.34 \pm 0.40$ & $3.96 \pm 0.52$ \\
\bottomrule
\end{tabular}
\end{table}

The limited variation in post-switch collision rates among
detection methods motivates our extended evaluation of adaptation
quality.

\begin{figure*}[t]
\centering
\includegraphics[width=0.8\textwidth]{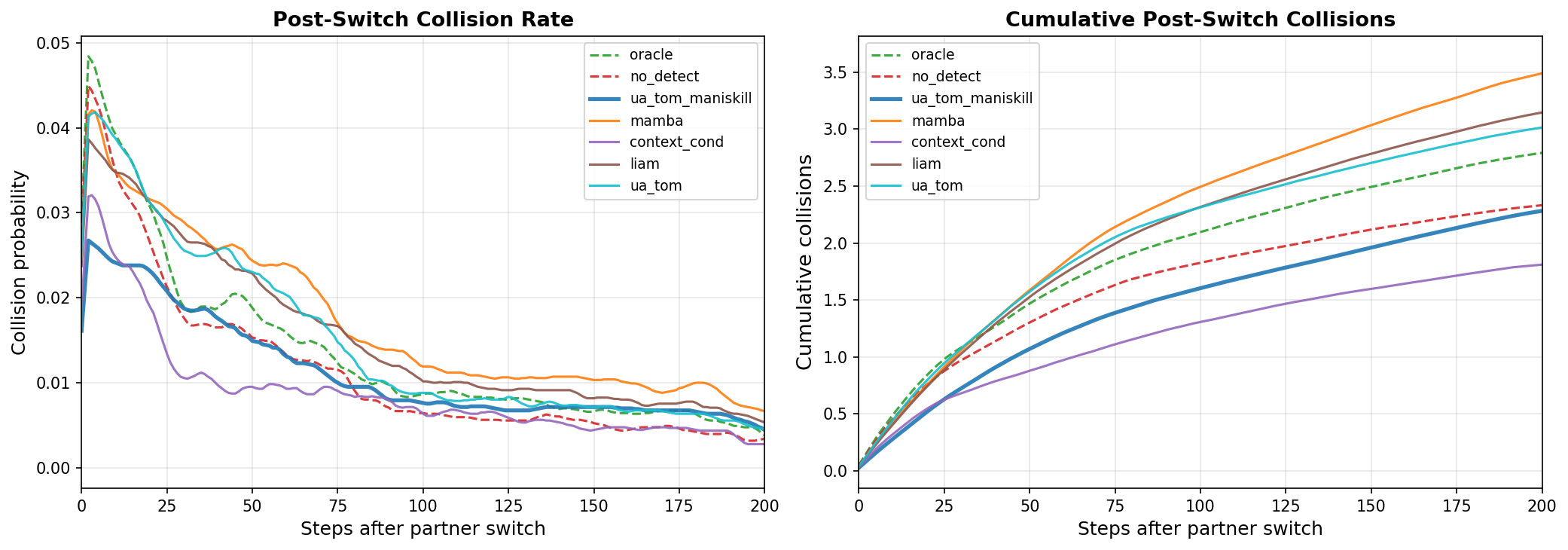}
\caption{Post-switch adaptation dynamics (300-step episodes, 10
seeds). \textbf{(A)}: Post-switch collision rate over time following
$t_s$, showing instantaneous risk at each timestep. \uatom{} (MS)
exhibits a smooth, monotonic decrease consistent with the SSM's
gradual belief revision (Fig.~\ref{fig:dynamics}). Oracle shows a
sharp transient from instantaneous strategy switching. LIAM remains
elevated for an extended window despite successful detection.
\textbf{(B)}: Cumulative post-switch collisions over time. The
separation between methods grows over the post-switch window,
confirming that sustained proximity behavior, not isolated collision
events, drives the safety differences captured by CRT.}
\label{fig:adaptation}
\end{figure*}

\textbf{Adaptation quality under extended episodes.}
Table~\ref{tab:adaptation} and Fig.~\ref{fig:adaptation} evaluate
adaptation quality using 300-step episodes and ten seeds. At this
episode length, detection rates saturate ($>$95\% for all methods),
shifting evaluation to adaptation quality.
We report two \uatom{} variants: \uatom{} (MS), trained and
evaluated on ManiSkill, and \uatom{}, trained on Overcooked and
evaluated zero-shot on ManiSkill. The former tests in-domain
adaptation quality; the latter tests cross-domain transfer.

\begin{table}[t]
\centering
\caption{Extended evaluation (300 steps, 10 seeds). Detection
saturates; the differentiator is adaptation quality. CRT =
close-range time (steps at $<$15\,cm post-switch).
$^\dagger$Privileged.}
\label{tab:adaptation}
\setlength{\tabcolsep}{3pt}
\scriptsize
\begin{tabular}{@{}lcccc@{}}
\toprule
\textbf{Method} & \textbf{Det\%}$\uparrow$ & \textbf{Post-Coll.}$\downarrow$ & \textbf{CRT}$\downarrow$ & \textbf{Total Coll.}$\downarrow$ \\
\midrule
Oracle & $100.0$ & $2.6 \pm 0.7$ & $5.3 \pm 0.9$ & $7.1 \pm 1.0$ \\
\midrule
Ctx-Cond.$^\dagger$ & $98.6 \pm 0.4$ & $\mathbf{2.1 \pm 0.6}$ & $4.9 \pm 1.0$ & $\mathbf{5.2 \pm 1.3}$ \\
\textbf{\uatom{} (MS)} & $97.5 \pm 0.9$ & $2.6 \pm 0.7$ & $\mathbf{4.8 \pm 1.1}$ & $5.9 \pm 1.4$ \\
Mamba & $98.5 \pm 0.4$ & $3.5 \pm 0.9$ & $7.0 \pm 0.7$ & $7.0 \pm 1.4$ \\
LIAM & $97.3 \pm 0.7$ & $3.4 \pm 0.5$ & $13.4 \pm 1.5$ & $7.1 \pm 0.9$ \\
\uatom{} & $95.1 \pm 0.8$ & $2.7 \pm 0.6$ & $9.2 \pm 1.8$ & $6.4 \pm 1.1$ \\
\midrule
No Detection & $0.0$ & $2.4 \pm 0.5$ & $5.0 \pm 0.9$ & $6.7 \pm 0.9$ \\
\bottomrule
\end{tabular}
\end{table}

\begin{figure*}[t]
\centering
\includegraphics[width=0.8\textwidth]{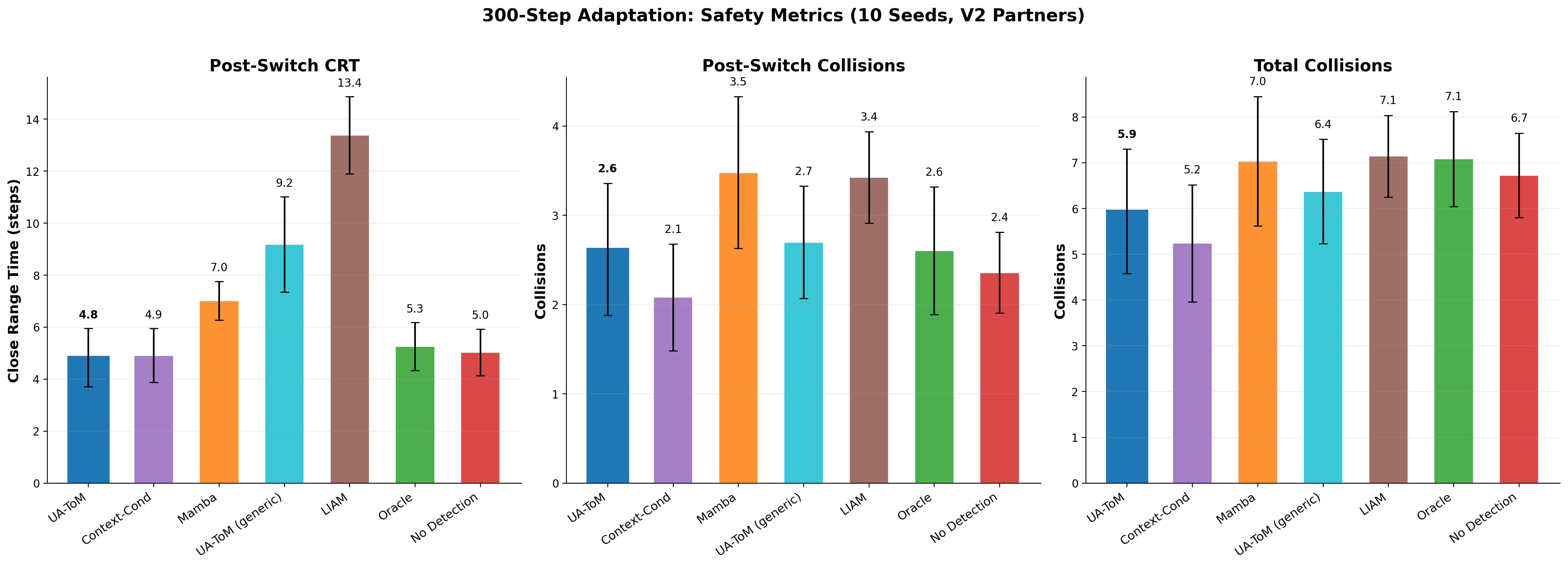}
\caption{Close-range time (CRT) under extended episodes (300 steps,
10 seeds). CRT measures post-switch timesteps at $<$15\,cm
end-effector distance. \uatom{} (MS) achieves the lowest CRT (4.8),
including lower than Oracle (5.3), indicating that smooth belief
revision maintains safer clearance than instantaneous strategy
switching. LIAM detects at 97.3\% yet accumulates $2.8\times$ more
danger time (13.4 steps). $^\dagger$Privileged.}
\label{fig:crt}
\end{figure*}

Three findings emerge. First, \uatom{} (ManiSkill variant) achieves
the lowest CRT at 4.8 steps (Fig.~\ref{fig:crt}), indicating it
maintains the safest clearance during post-switch adaptation. This
is lower than the Oracle (5.3), which has perfect detection but
produces a trajectory discontinuity at the switch point that
temporarily enters the danger zone. Second, detection rate does not
predict adaptation quality. LIAM achieves 97.3\% detection but
accumulates 13.4 CRT steps---nearly three times higher than
\uatom{}. Third, Context-Conditioned achieves the lowest collision
count (2.1) through access to ground-truth type labels; among
unassisted methods, \uatom{} (MS) provides the best safety profile
on CRT.

\begin{figure}[t]
\centering
\includegraphics[width=0.85\columnwidth]{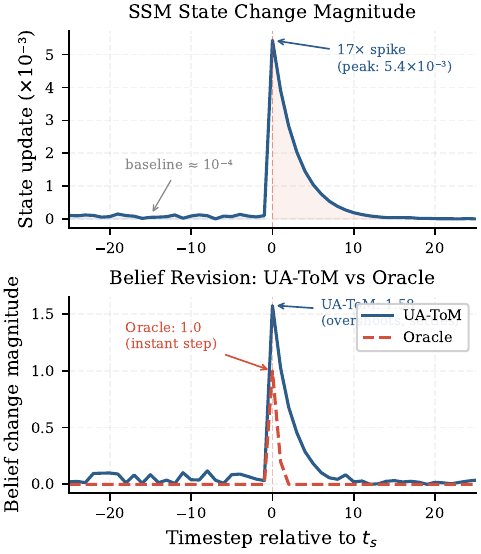}
\caption{Internal dynamics around the regime switch (24 episodes,
ManiSkill). \textbf{Top}: SSM state update magnitude
$\|\mathbf{b}_t - \mathbf{b}_{t-1}\|$ spikes $17\times$ at $t_s$
(from $9.4 \times 10^{-5}$ to $5.4 \times 10^{-3}$) and decays
over roughly 10 timesteps. \textbf{Bottom}: \uatom{} belief change
overshoots the Oracle step function (1.58 vs.\ 1.0) then settles
gradually over 5--10 steps. The SSM discretization $\Delta_t$
remains near-constant at ${\approx}0.78$ throughout, confirming
that detection sensitivity resides in the learned dynamics matrices
$\mathbf{A}, \mathbf{B}$ rather than input-dependent gating.}
\label{fig:dynamics}
\end{figure}

\textbf{Adaptation dynamics.} Fig.~\ref{fig:adaptation} shows the
temporal profile of post-switch adaptation. \uatom{}'s collision
risk decreases smoothly after the switch, consistent with the
gradual belief revision observed in the SSM dynamics
(Fig.~\ref{fig:dynamics}). Methods with higher CRT show elevated
risk for longer windows after the switch.

\textbf{Latency.} \uatom{} detects regime changes in 9.9 steps on
average (Table~\ref{tab:full}), the fastest among practical
detectors. GRU requires over 12 steps and exhibits the largest
variance. \uatom{}'s median latency of 2 steps (P75 = 3) explains
its strong performance under $\pm 3$ tolerance.

\textbf{Cross-domain reliability (Overcooked).}
Table~\ref{tab:overcooked} evaluates detection reliability in
Overcooked across two layouts and two protocol variants. The trend
observed in ManiSkill persists. \uatom{} consistently achieves the
highest detection floor and the lowest coefficient of variation
across all settings. The most challenging configuration (V1 Cramped)
shows a 30 percentage point gap: \uatom{} at 87\% versus GRU at 57\%.

\begin{table}[t]
\centering
\caption{Detection reliability across Overcooked layouts. Floor =
worst-seed detection rate ($\mu - \sigma$). CV = coefficient of
variation across five seeds.}
\label{tab:overcooked}
\setlength{\tabcolsep}{2.5pt}
\scriptsize
\begin{tabular}{@{}l|cc|cc|cc|cc@{}}
\toprule
& \multicolumn{2}{c|}{\textbf{Cramped-1}} & \multicolumn{2}{c|}{\textbf{Asym.-1}} & \multicolumn{2}{c|}{\textbf{Cramped-2}} & \multicolumn{2}{c}{\textbf{Asym.-2}} \\
\textbf{Method} & Fl. & CV & Fl. & CV & Fl. & CV & Fl. & CV \\
\midrule
\textbf{\uatom{}} & \textbf{87} & \textbf{.048} & \textbf{99} & \textbf{.007} & \textbf{99} & \textbf{.005} & \textbf{97} & \textbf{.017} \\
Transformer & 83 & .049 & 99 & .006 & 97 & .013 & 97 & .013 \\
GRU & 57 & .208 & 86 & .061 & 92 & .033 & 90 & .033 \\
\bottomrule
\end{tabular}
\end{table}

\textbf{Representation separability.}
Table~\ref{tab:separability} analyzes embedding separability using
Bhattacharyya distance and silhouette score. BOCPD achieves the
largest mean Bhattacharyya distance (322.2) with a reasonable
silhouette score (0.40), indicating strong type separation. Its 0\%
open-loop switch F1 therefore arises from an architectural limitation
rather than representational weakness. BToM exhibits genuine
representational collapse (silhouette 0.17). \uatom{}'s contrastive
memory achieves the best type-discriminative representation
(silhouette 0.52).

\begin{table}[t]
\centering
\caption{Representation analysis. BOCPD has the highest mean
separation yet 0\% switch F1. CM = contrastive memory.}
\label{tab:separability}
\setlength{\tabcolsep}{3pt}
\scriptsize
\begin{tabular}{@{}lcccc@{}}
\toprule
\textbf{Model} & \textbf{Min Dist.} & \textbf{Mean Dist.} & \textbf{Silhouette} & \textbf{Hardest Pair} \\
\midrule
BOCPD & 38.1 & \textbf{322.2} & 0.40 & Help.--Comp. \\
GRU & 34.5 & 127.5 & 0.40 & Help.--Comp. \\
LIAM & 22.5 & 89.0 & 0.47 & Help.--Comp. \\
Ctx-Cond. & 31.5 & 77.1 & 0.38 & Help.--Block. \\
Mamba & 33.8 & 76.2 & 0.38 & Help.--Block. \\
\uatom{} & 31.9 & 74.6 & 0.39 & Help.--Block. \\
Transformer & 29.9 & 68.6 & 0.38 & Help.--Block. \\
BToM & 12.8 & 16.7 & 0.17 & Help.--Comp. \\
\midrule
\uatom{} + CM & \textbf{41.8} & 146.2 & \textbf{0.52} & --- \\
\bottomrule
\end{tabular}
\end{table}

\textbf{Internal dynamics.} Fig.~\ref{fig:dynamics} visualizes
\uatom{}'s internal dynamics around the regime switch, averaged
over 24 episodes. Three observations explain the detection behavior.
First, the SSM discretization step $\Delta_t$ remains nearly
constant at approximately 0.78 throughout the episode, including
across the switch. Second, the hidden-state update magnitude
$\|\mathbf{b}_t - \mathbf{b}_{t-1}\|$ exhibits a sharp spike at
the switch point, increasing roughly $17\times$ relative to baseline
(from $9.4 \times 10^{-5}$ to $5.4 \times 10^{-3}$) and decaying
over approximately ten timesteps. Third, the belief update
temporarily overshoots the oracle signal (1.58 compared with 1.0)
before settling over 5--10 steps. Action prediction error increases
$2.3\times$ at $t_s + 1$, providing the perturbation that triggers
the SSM update cascade. This gradual revision is the mechanism
underlying \uatom{}'s low CRT: smooth trajectory adaptation
maintains end-effector clearance during the transition
(Fig.~\ref{fig:adaptation}).

\textbf{Cross-domain ablation.}
Table~\ref{tab:ablation} evaluates the contribution of individual
components. The dominant component differs by domain. In continuous
manipulation (ManiSkill), hierarchical prediction error is critical:
removing it reduces switch F1 from 0.558 to 0.121 ($-$78\%). In
discrete coordination (Overcooked), causal attention dominates;
removing it increases detection latency from 13.3 to 26.4 steps
(+13.1). This asymmetry reflects differences in observation
structure.

\begin{table}[t]
\centering
\caption{Ablation: ManiSkill (open-loop F1) and Overcooked
(closed-loop latency, Cramped Room).}
\label{tab:ablation}
\setlength{\tabcolsep}{3.5pt}
\small
\begin{tabular}{@{}lcc|cc@{}}
\toprule
& \multicolumn{2}{c|}{\textbf{ManiSkill}} & \multicolumn{2}{c}{\textbf{Overcooked}} \\
\textbf{Configuration} & F1 & $\Delta$ & Lat. & $\Delta$ \\
\midrule
Full \uatom{} & 0.558 & --- & 13.3 & --- \\
$-$ Hier.\ Pred.\ Error & 0.121 & $-$78\% & 14.9 & +1.6 \\
$-$ Causal Attention & 0.412 & $-$26\% & 26.4 & +13.1 \\
$-$ Selective SSM & 0.479 & $-$14\% & --- & --- \\
$-$ Contrastive Memory & 0.502 & $-$10\% & 17.5 & +4.2 \\
$-$ Integration Layer & 0.531 & $-$5\% & 13.6 & +0.3 \\
\bottomrule
\end{tabular}
\end{table}

The ManiSkill ablation is single-seed; we report it as directional
evidence consistent with the Overcooked results.

\textbf{Training modality.} We train \uatom{} on top of a frozen
backbone, which achieves 85.7\% switch detection in 6.5 minutes on
an RTX 6000 Ada. Freezing the backbone is sufficient for the
detection task and avoids modifying the base control policy.

\textbf{Inference overhead.}
Table~\ref{tab:inference} reports inference latency. \uatom{}
requires 7.41\,ms per forward pass (14.8\% of a 50\,ms budget).
The SSM sequential scan accounts for 64.4\% of latency; the
attention path costs only 10.9\% due to batched operations.

\begin{table}[t]
\centering
\caption{Inference latency ($T{=}50$ window, RTX 6000 Ada). All
methods fit within the 50\,ms control budget.}
\label{tab:inference}
\setlength{\tabcolsep}{3.5pt}
\small
\begin{tabular}{@{}lrcrc@{}}
\toprule
\textbf{Method} & \textbf{Params} & \textbf{Mean (ms)} & \textbf{P99 (ms)} & \textbf{Budget} \\
\midrule
GRU & 119K & 0.40 & 1.94 & 0.8\% \\
LIAM & 215K & 0.62 & 2.16 & 1.2\% \\
Transformer & 496K & 0.97 & 2.78 & 1.9\% \\
\textbf{\uatom{}} & \textbf{992K} & \textbf{7.41} & \textbf{10.36} & \textbf{14.8\%} \\
Ctx-Cond. & 259K & 8.54 & 11.10 & 17.1\% \\
Mamba & 261K & 9.27 & 11.50 & 18.5\% \\
BOCPD & 226K & 19.36 & 28.08 & 38.7\% \\
BToM & 232K & 24.10 & 27.75 & 48.2\% \\
\bottomrule
\end{tabular}
\end{table}

\section{Discussion}

Two observations emerge from our experiments. \textit{First}, the
mechanistic analysis indicates that regime detection arises from the
dynamics of the selective state-space update rather than from
input-dependent gating. Around the switch point the hidden-state
update magnitude increases by roughly $17\times$ while the
discretization step remains nearly constant. This produces a stable
spike--decay signal driven by prediction-error perturbations,
allowing the detector to remain insensitive to steady-state
variation while reacting strongly to behavioral regime changes. The
same dynamics produce smooth belief revision that maintains
end-effector clearance during adaptation, resulting in lower
close-range time than instantaneous strategy switching.
\textit{Second}, the ablation results show that the dominant
architectural component depends on the task structure. In continuous
manipulation, prediction-error signals provide the primary detection
cue, whereas in discrete coordination tasks attention-based temporal
aggregation becomes more important. This suggests that regime-switch
detection benefits from modular designs where temporal evidence
accumulation and prediction-error signals can contribute with
different weights depending on the observation structure. Our
evaluation is limited to scripted partner policies with discrete
types and state-based observations; extending to continuous human
behavior and vision-based inputs remains future work.

\section{Conclusion}
\label{sec:conclusion}

We evaluated ten regime-switch detection methods across five random
seeds in shared-workspace manipulation. Two findings emerge. First,
enabling switch detection reduces post-switch collisions by 52\%,
yet existing methods show large reliability variation under
operationally realistic tolerance, with detection rates ranging from
86\% to 30\% at $\pm 3$ steps. Such differences remain hidden under
standard $\pm 5$ evaluation. Second, detection rate alone does not
determine safety outcomes. Methods with comparable detection
($>$97\%) produce close-range times varying from 4.8 to 13.4 steps.
\uatom{}, a 992K-parameter belief-tracking module, achieves the
highest detection rate among unassisted methods and the lowest
close-range time, including lower than an Oracle with perfect
detection. Mechanistic analysis shows that robustness arises from
selective state-space dynamics. Hidden-state updates increase
$17\times$ at regime switches while the discretization step remains
nearly constant, producing a stable change signal across observation
conditions. Cross-domain ablations further show that architectural
contributions depend on task structure, supporting a modular design
for regime-switch detection under collaboration.


\end{document}